\def\year{2022}\relax
\documentclass[letterpaper]{article} % DO NOT CHANGE THIS
\usepackage{aaai22}  % DO NOT CHANGE THIS
\usepackage{times}  % DO NOT CHANGE THIS
\usepackage{helvet}  % DO NOT CHANGE THIS
\usepackage{courier}  % DO NOT CHANGE THIS
\usepackage[hyphens]{url}  % DO NOT CHANGE THIS
\usepackage{graphicx} % DO NOT CHANGE THIS
\usepackage{natbib}  % DO NOT CHANGE THIS AND DO NOT ADD ANY OPTIONS TO IT
\usepackage{caption} % DO NOT CHANGE THIS AND DO NOT ADD ANY OPTIONS TO IT
\DeclareCaptionStyle{ruled}{labelfont=normalfont,labelsep=colon,strut=off} % DO NOT CHANGE THIS
\makeatletter
\def\orcidID#1{\smash{\href{https://orcid.org/#1}{\protect\raisebox{-1.25pt}{\protect\includegraphics{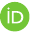}}}}}
\makeatother
\usepackage{algorithm}
\usepackage{algorithmic}

\usepackage{amsmath}
\usepackage{amsfonts}
\usepackage{mathtools}
\usepackage{booktabs}
\usepackage{multirow}

\newtheorem{definition}{Definition}

\DeclareMathOperator{\Clause}{PC} % Proposition Constraint
\DeclareMathOperator{\Actionclause}{AC} % Action clause
\DeclareMathOperator{\Uclause}{UPC} % Ungrounded Action clause
\DeclareMathOperator{\GT}{T_g} % Grounded transition function
\DeclareMathOperator{\GTi}{T_g^{\mathit i}} % Grounded transition function
\DeclareMathOperator{\UTi}{T_u^{\mathit i}} % Ungrounded transition function
\DeclareMathOperator{\RC}{RC} % restricted clauses

\DeclareMathOperator{\ground}{GD}
\DeclareMathOperator{\bin}{bin}
\DeclareMathOperator{\pre}{pre} % preconditions
\DeclareMathOperator{\eff}{eff} % effects
\DeclareMathOperator{\arity}{arity} % predicate arity

\DeclareMathOperator{\UPL}{\Pi} % Ungrounded planning problem
\DeclareMathOperator{\UI}{I_u} % Ungrounded initial state
\DeclareMathOperator{\Atom}{Atom} % predicates
\DeclareMathOperator{\operator}{A} % operators set
\DeclareMathOperator{\initial}{I} % initial state
\DeclareMathOperator{\goal}{G} % goal state
\DeclareMathOperator{\UG}{G_u} % Ungrounded goal state
\DeclareMathOperator{\PR}{P} % predicates
\DeclareMathOperator{\OBJ}{O} % objects
\DeclareMathOperator{\St}{F} % grounded state
\DeclareMathOperator{\OBJSVARS}{OC} % objects variable combinations
\DeclareMathOperator{\PARVARS}{PM} % parameter variables

\DeclareMathOperator{\onef}{f} % one grounded atom/fluent
\DeclareMathOperator{\oneg}{g} % goal state
\DeclareMathOperator{\oneq}{q} % predicates
\DeclareMathOperator{\onep}{p} % predicates
\DeclareMathOperator{\ob}{o} % single object
\DeclareMathOperator{\op}{a} % operator
\DeclareMathOperator{\onepm}{x} % single parameter
\DeclareMathOperator{\oneoc}{y} % one object combination

\DeclareMathOperator{\maxarg}{\eta} % maximum arguments for predicates
\DeclareMathOperator{\maxpar}{\zeta} % maximum arguments for parameters
\DeclareMathOperator{\numvar}{\gamma} % effects
\DeclareMathOperator{\opvar}{\sigma} % parameter index
\DeclareMathOperator{\numoc}{\delta} % number of object combinations
\DeclarePairedDelimiter\ceil{\lceil}{\rceil}

\DeclareMathOperator{\unstack}{unstack} % unstack action
\DeclareMathOperator{\stack}{stack} % stack action
\DeclareMathOperator{\clear}{clear} % predicate
\DeclareMathOperator{\ontable}{ontable} % predicate
\DeclareMathOperator{\on}{on} % predicate

\usepackage{newfloat}
\usepackage{listings}

\lstdefinelanguage{stripspddl}
{
  sensitive=false,    % not case-sensitive
  morecomment=[l]{;}, % line comment
  alsoletter={:},   % consider extra characters
  morekeywords={
    define,domain,problem,not,and,or,
    :domain,:requirements,:types,:objects,:constants,
    :predicates,:action,:parameters,:precondition,:effect,
    :init,:goal,:strips,:equality,:typing,:negative-preconditions
  }
}

\floatstyle{ruled}
\newfloat{listing}{tb}{lst}{}
\floatname{listing}{Listing}
\usepackage{hyperref}

\title{Classical Planning as QBF Without Grounding \\ (extended version)}
\author {
    Irfansha Shaik\textsuperscript{\rm 1} \orcidID{0000-0002-7404-348X},
    Jaco van de Pol\textsuperscript{\rm 1} \orcidID{0000-0003-4305-0625}
}
\affiliations {
    \textsuperscript{\rm 1} Aarhus University, Department of Computer Science, Aarhus, Denmark\\
}
\begin{document}

\nocopyright

\maketitle

\begin{abstract}
Most classical planners use grounding as a preprocessing step, essentially
reducing planning to propositional logic. However, grounding 
involves instantiating all action rules with concrete object combinations, and
results in large encodings for SAT/QBF-based planners.
This severe cost in memory becomes a main bottleneck when actions have many parameters,
such as in the Organic Synthesis problems from the IPC 2018
competition.
We provide a compact QBF encoding that is logarithmic in the number of objects and avoids grounding completely, 
by using universal quantification for object combinations.
We show that we can solve some of the Organic Synthesis problems, which could not be
handled before by any SAT/QBF based planners due to grounding.
\end{abstract}

\section{Introduction}
\label{sec:introduction}

Automated planning has many real-world applications, such as Space Exploration and Robotics, cf.\ the book by \citet{ghallab2004automated}.
The two main research directions in automated planning are
heuristic-based state-space search and propositional satisfiability (SAT) based solving.
While heuristic-based search often finds some plan quickly, it may not guarantee to search the whole search space.
SAT-based solvers, on the other hand, can also be used to prove the non-existence of plans up to a bounded length.
For some applications, quickly falsifying the existence of a plan can be useful.
Classical planning is the most simple problem; its aim is to find a valid sequence of actions from a single initial state to some goal state, where the state is completely known and the effect of all
actions is deterministic.
\citet{Kautz1992planning} reduced the planning problem to the bounded reachability problem and provided a corresponding SAT encoding for some plan length $k$.

Classical planning domains are usually defined by logical rules that describe the pre-conditions and effects of applying actions.
Usually, a single action is applied to multiple objects, corresponding to the arity of that action. 
The pre-conditions and effects contain predicates, which also refer to multiple objects.
A concrete planning problem defines a universe of concrete objects, and an initial and goal condition. 
Many planning tools, both SAT-based and heuristic,
first apply a grounding step: all action rules are instantiated for all possible combinations of objects. However, 
for domains with actions that involve many objects, the grounded specification is very large, 
and memory becomes the main bottleneck.
Despite several improvements, such as action splitting \cite{Kautz1992planning}, explanatory frame axioms \cite{haas1987case}, invariants \cite{rintanen2008regression} and parallel plans \cite{RINTANEN20061031}, the memory  still remains a bottleneck for domains with large action arity.

QBF (Quantified Boolean Formula) encodings are known for being more compact than SAT encodings,
so they are considered as an alternative when SAT encodings suffer from memory problems.
\citet{DBLP:conf/sat/DershowitzHK05} and \citet{JUSSILA200745} proposed a QBF encoding with $\exists\forall\exists$ quantifier alternation for reachability in Bounded Model Checking (BMC). It generates only a single copy of the transition function, instead of $k$ copies in SAT, by using quantification over state variables.
\citet{DBLP:conf/lpar/Rintanen01} proposed a reachability QBF encoding that is logarithmic in the length of the plan, and uses only one transition function.
\citet{cashmore2012planning} proposed the Compact Tree Encoding (CTE), which improves upon the logarithmic encoding by \citeauthor{DBLP:conf/lpar/Rintanen01} (in the context of planning) by efficient traversal of the search tree.
Although these QBF encodings are more concise than the SAT encodings, it has been reported that the
SAT encodings can usually be solved faster than the QBF encodings by the current solvers \cite{cashmore2012planning}.
More importantly, the QBF encodings mentioned above still require the problem to be grounded first, 
so the memory bottleneck for actions that involve many objects has not been solved.

\citet{Matloob2016ExploringOS} proposed Organic Synthesis benchmarks and showed that SAT based encodings could not solve any of the non-split instances.
They compared Madagascar (among other SAT based planners) and Fast Downward \cite{DBLP:journals/corr/abs-1109-6051} on these benchmarks.
This is consistent with the findings from the IPC-$2018$ planning competition, 
in which SAT based planners performed poorly on instances from Organic Synthesis. 
These benchmarks contain actions that manipulate up to 17 objects,
so the grounding step exhausts the memory. This not only applies to all SAT/QBF based planners
based on grounding:
Even heuristic planners that employ grounding will exhaust the memory for this domain.
Thus, to solve these problems, there is a need for encodings that avoid grounding altogether.

\paragraph{Our Contribution and Overview of the Paper.}
In this paper, we propose a QBF encoding of quantifier structure $\exists\forall\exists$ which completely avoids grounding, by using universal variables to represent combinations of objects. The encoding is linear in the number of action names, predicates, and path length, and logarithmic in the number of objects.
We provide an open-source implementation
\footnote{https://github.com/irfansha/Q-Planner} 
of the encoding.

We will explain our encoding in two steps, to make it more accessible. In the first step (Section~\ref{sec:planningassat}), we present
an intermediate representation in SAT. The difference with the traditional binary encoding is that we provide a little more structure:
To encode a grounded action $A(o_1,\ldots,o_n)$, we use a binary encoding of the action name $A$, followed
by binary encodings of each object $o_i$. The number of bits is logarithmic in the number of action names
and the number of possible objects. We organize the constraints as a big conjunction over all atoms, i.e., predicates
instantiated with concrete object combinations.
We do not claim that this SAT encoding is more efficient than the traditional SAT encoding. Instead, this extra structure
just makes the transition to the QBF encoding more smooth.

The SAT encoding still contains constraints for each grounded predicate and action. Hence the size of the encoding
grows exponentially in the number of parameters of predicates and actions. 
Our QBF encoding (Section~\ref{sec:ungroundedqbfencoding}) avoids this as follows: 
Each constraint is encoded only once, for a ``symbolic'' object combination $(o_1,\ldots,o_n)$. We use universal quantifiers
over these variables to have them instantiated for each concrete object combination, rather than explicit grounding.
Now the constraints are organized as a conjunction over ungrouded predicate names only, and the number of
variables is logarithmic in the number of objects.

In our experiments (Section~\ref{sec:analysis}), we generated the QBF encoding for 20 domains from 
previous IPC competitions, and solve them with the QBF solver CAQE. 
We compare the size of our QBF encoding with the simple SAT encoding of the best SAT-based planner,
Madagascar. We also compare the running time of solving the QBF and SAT encodings.

Focusing on the Organic Synthesis benchmarks, we compare our proposal
not only with Madagascar (using its best encoding), but also with two of the best heuristic solvers.
We show that some of the Organic Synthesis problems that could not be handled before by any SAT/QBF based planners can now be solved using our encoding. In these cases, we can also show that there are no plans of shorter length.

\section{Preliminaries}
\label{sec:preliminaries}

We introduce the main notions of the classical planning problems and QBF.
We also present a running example of the blocks-world domain (Listing~\ref{lst:blocksdomain})
in PDDL, the Planning Domain Definition Language \cite{DBLP:journals/jair/FoxL03}.

\subsection{Classical Planning}
\label{subsec:classicalplanning}

\begin{lstlisting}[mathescape,
  float=!b,
  caption={blocks-world domain and problem.},
  label={lst:blocksdomain},
  language=stripspddl,numbers=none]
(define (domain blocksworld)
(:predicates (clear ?$y_1$)(ontable ?$y_1$)
			(on ?$y_1$ ?$y_2$))
(:action unstack
 :parameters (?$x_1$ ?$x_2$)
 :precondition (and (clear ?$x_1$)
 			(on ?$x_1$ ?$x_2$))
 :effect (and (not (on ?$x_1$ ?$x_2$))
 			(ontable ?$x_1$)(clear ?$x_2$)))
(:action stack
 :parameters (?$x_1$ ?$x_2$)
 :precondition (and (clear ?$x_1$)
 			(clear ?$x_2$)(ontable ?$x_1$))
 :effect (and (not(clear ?$x_2$))
 			(not(ontable ?$x_1$))(on ?$x_1$ ?$x_2$))))
(define (problem BW_rand_2)
(:domain blocksworld)
(:objects $b_1$ $b_2$)
(:init (ontable $b_1$) (on $b_2$ $b_1$) (clear $b_2$))
(:goal (and (on $b_1$ $b_2$))))
\end{lstlisting}

\noindent We start with introducing a planning signature.

\begin{definition}
\label{def:planningsignature}
A planning signature $\Sigma = \langle \PR, \operator, \OBJ, \arity \rangle$
is a $4$-tuple where $\PR$ is a set of predicate symbols, $\operator$ is a set of action names, $\OBJ$ is a set of objects, and
the function $\arity : (\PR \cup \operator \to  \mathbb{N}_0)$ indicates the expected number of arguments.
\end{definition}

\noindent
In the example of Listing~\ref{lst:blocksdomain}, 
$\PR = \{\clear, \ontable, \on\}$, $\OBJ = \{b_1, b_2\}$, $\operator =  \{\unstack, \stack\}$,
and $\arity(\on)=2$.
A schematic action corresponds to an action name with parameters, e.g. $\stack(?x_1,?x_2)$. It can be
grounded to a set of concrete actions, by replacing the parameters by concrete object combinations.
Predicate symbols are also equipped with parameters, that can be grounded by concrete objects.
To facilitate our encodings, we unify the parameter names of all actions to a fixed sequence of $x$'s, and the 
formal predicate parameters to a sequence of $y$'s.
Note that these action and predicate parameter names will be used later for (universal) object combination variables in the QBF encoding.
Since we use a sequential plan semantics, a single set of action and predicate parameter variables is sufficient to represent any chosen action at each time step.

Each action is specified schematically by preconditions and effects, which we will discussed next.

\begin{definition}
\label{def:atoms}
Given a signature, we define the set of atoms and grounded atoms (or fluents) as:
\begin{align*}
\Atom &= \{ \onep(\overrightarrow{x}) \mid \onep\in \PR, |\overrightarrow{x}|=\arity(\onep)\}\\
\St &= \{ \onep(\overrightarrow{\ob}) \mid \onep\in \PR, \ob_i\in\OBJ, |\overrightarrow{\ob}|=\arity(\onep) \}
\end{align*}
\end{definition}

\begin{definition}
\label{def:ungroundedplanningproblem}
Given a signature $\Sigma = \langle \PR, \operator, \OBJ, \arity \rangle$.
The planning problem $\UPL = \langle \initial, \goal, \pre^+, \pre^-, \eff^+, \eff^- \rangle$ is a $6$-tuple where
\begin{itemize}
\item Initial state $\initial \subseteq \St$ and
\item Goal condition $\goal = (g^{+}, g^{-})$, where $g^{+}, g^{-} \subseteq \St$.
\item for each action $\op \in \operator$, we have:
  \begin{itemize}
  \item positive preconditions $\pre^{+}(\op) \subseteq \Atom$
  \item negative preconditions $\pre^{-}(\op) \subseteq \Atom$
  \item positive effects $\eff^{+}(\op) \subseteq \Atom$ 
  \item negative effects $\eff^{-}(\op) \subseteq \Atom$
  \end{itemize}
\end{itemize}
\end{definition}
In the example, (on ?$x_1$ ?$x_2$) is a positive precondition for
action \texttt{unstack}, while (ontable ?$x_1$) is a negative effect for action \texttt{stack}.

\subsection{SAT encoding}
\label{subsec:sat}

The planning problem is encoded as finding a path in a graph, whose nodes
consist of assignments to the fluents and whose edges are defined by the conditions and effects of grounded actions.
Finding a plan of length $k$ corresponds to finding a path of length $k$ 
from the intial node to a goal node.
Usually, a SAT encoding uses propositional variables encoding the $k$ actions and $k+1$ states in the path.
A direct encoding would use $k$ variables for each grounded action, but since we use sequential plan semantics 
(in which only one action occurs at the time), this can be encoded by a number of variables that is logarithmic in the number of grounded
actions (still linear in $k$).

We will present the constraints in circuit format (with nested $\wedge$ and $\vee$), which can be transformed
to CNF clauses by a standard procedure. We will write $\bigwedge$ and $\bigvee$ for conjunctions and
disjunctions over finite (indexed) sets.

\subsection{Quantified Boolean Formulas}
\label{subsec:qbf}

Quantified Boolean Formulas (QBF) extend propositional formulas in SAT with universal quantification over some variables.
A QBF formula in \emph{prenex normal form} is of the shape $\Pi\phi$ where $\phi$ is called the matrix, which is a propositional formula, and $\Pi = Q_1 X_1 \dots Q_n X_n$ is the prefix with alternating existential and universal quantifiers $Q_i\in\{\forall,\exists\}$ and disjoint sets of variables $X_i$. The $Q_i$ specify the quantification of each variable that occurs in $\phi$. We consider only closed formulas, i.e., the quantification of all variables is known.
A QBF is evaluated to either True or False and its truth value can be computed by recursively solving the formula over each outermost variable. A formula $\exists x \phi$ is true if and only if $\phi[\top/x]$ or $\phi[\bot/x]$ is true. A formula $\forall x \phi$ is true if and only if both $\phi[\top/x]$ and $\phi[\bot/x]$ are true.
The order within each quantified block does not influence the truth value of the formula.
When expanded, a QBF corresponds to an and-or tree. 

\section{Intermediate SAT encoding}
\label{sec:planningassat}

We now provide the intermediate SAT encoding of planning problems, as an introduction to our QBF encoding. Throughout the section, we assume a fixed planning problem as in Definition~\ref{def:ungroundedplanningproblem}. Recall that we introduced a fixed
sequence of action parameters $x$.

For the encoding, we generate copies of variables for each time step (represented by superscript) for a path of length $k$.
We use action variables $\operator^{i} = \{\op^{i}_{b}\mid 1 \leq b \leq \opvar \}$ for each $0\leq i \leq k-1$ and $\opvar = \ceil{\log(|\operator|)}$; here $\op_b^i$ represents the $b$-th bit of a logarithmic encoding of the action name that occurs at time $i$ in the plan.
The action parameter variables are $\PARVARS^i = \{ \onepm^i_{j, b} \mid 1 \leq j \leq \maxpar, 1\leq b\leq \numvar \}$ for each
$0\leq i \leq k-1$, where $\maxpar$ is the maximum action arity and $\numvar = \ceil{\log(|\OBJ|)}$.
Here $\onepm_{j,b}^i$ represents the $b$-th bit of parameter $\onepm_j$ of the action scheduled at time $i$ in the plan. We will write 
$\overrightarrow{\onepm^i_j}$ (sequence of $\numvar$ variables) for the parameter $\onepm_j$ at time $i$.
The state variables are $\St^i = \{ \onef^i_{\onep(\overrightarrow{\ob})} \mid \onep(\overrightarrow{\ob})\in \St\}$, i.e., one variable for every fluent and for each $0\leq i \leq k$.

The corresponding SAT encoding is based on these variables, and constraints for the initial condition $\initial$,
goal condition $\goal$, transition function $\GT$, and a domain restriction $\RC$.

\begin{align*}
&\exists \operator^0, \PARVARS^0, \dots, \operator^{k-1}, \PARVARS^{k-1} \exists {\St^0, \ldots, \St^k} \\
&\initial(\St^0) \land \goal(\St^k) \land \bigwedge_{i=0}^{k-1}\GTi(\St^i, \St^{i+1},  \operator^i, \PARVARS^i) \land {} \\
&\bigwedge_{i=0}^{k-1} \RC(\operator^i, \PARVARS^i)
\end{align*}

\noindent In the constraint for the initial state, each variable is positive if the proposition is in the set $\initial$ and negative if it is not. For the goal state, the variable is positive if the proposition is in $\oneg^{+}$ and negative if it is in $\oneg^{-}$. Note that this provides a unique initial state, but
there may be multiple states that satisfy the goal condition.
We group the constraints (throughout the artice) based on the predicates for the sake of easy understanding of the correctness proof (in Section \ref{sec:correctness}).

\begin{definition}
\label{def:initialstate}
Initial constraint $\initial(\St^0) =$
\begin{align*}
&\bigwedge_{\onep \in \PR}\bigwedge_{\onep(\overrightarrow{\ob}) \in \St} \{\onef^{0}_{\onep(\overrightarrow{\ob})} | \onep(\overrightarrow{\ob}) \in \initial\} \land {} \\ 
&\bigwedge_{\onep \in \PR}\bigwedge_{\onep(\overrightarrow{\ob}) \in \St} \{\neg \onef^{0}_{\onep(\overrightarrow{\ob})} | \onep(\overrightarrow{\ob}) \notin \initial\}
\end{align*}
\end{definition}
\begin{definition}
Goal constraint $\goal(\St^k) =$
\begin{align*}
&\bigwedge_{\onep \in \PR}\bigwedge_{\onep(\overrightarrow{\ob}) \in \St} \{\onef^{k}_{\onep(\overrightarrow{\ob})} | \onep(\overrightarrow{\ob}) \in \oneg^{+}\} \land {}\\ 
&\bigwedge_{\onep \in \PR}\bigwedge_{\onep(\overrightarrow{\ob}) \in \St} \{\neg \onef^{k}_{\onep(\overrightarrow{\ob})} | \onep(\overrightarrow{\ob}) \in \oneg^{-}\}
\end{align*}
\end{definition}

To encode the transition function, we will generate five constraints for each proposition variable. 
These constraints define the value of all fluents and time stamps,
just before or after some action occurs.
The first four constraints correspond to the positive preconditions, negative preconditions, 
positive effects and negative effects, respectively. The last constraint corresponds to
the frame axiom, which indicates that untouched propositions should not change.

\begin{definition}
\label{def:groundedtf}
The \textbf{grounded transition function} is:
\begin{displaymath}
\GTi(\St^i, \St^{i+1},  \operator^i, \PARVARS^i) = \bigwedge_{\onep \in \PR}\bigwedge_{\onep(\overrightarrow{\ob}) \in \St} \Clause_{\onep(\overrightarrow{\ob})}^i
\end{displaymath} where
the \textbf{Proposition Constraint} $\Clause_{\onep(\overrightarrow{\ob})}^i$ =
\begin{align*}
&(\Actionclause_{\onep(\overrightarrow{\ob})}(\operator^i, \PARVARS^i, \pre^{+}) \implies {\onef}^i_{\onep(\overrightarrow{\ob})}) \wedge {}\\
&(\Actionclause_{\onep(\overrightarrow{\ob})}(\operator^i, \PARVARS^i, \pre^{-}) \implies \neg {\onef}^{i}_{\onep(\overrightarrow{\ob})}) \wedge {}\\
&(\Actionclause_{\onep(\overrightarrow{\ob})}(\operator^i, \PARVARS^i, \eff^{+}) \implies {\onef}^{i+1}_{\onep(\overrightarrow{\ob})}) \wedge {}\\ 
&(\Actionclause_{\onep(\overrightarrow{\ob})}(\operator^i, \PARVARS^i, \eff^{-}) \implies \neg {\onef}^{i+1}_{\onep(\overrightarrow{\ob})}) \wedge {}\\
&(({\onef}^i_{\onep(\overrightarrow{\ob})} = {\onef}^{i+1}_{\onep(\overrightarrow{\ob})}) \vee \Actionclause_{\onep(\overrightarrow{\ob})}(\operator^i, \PARVARS^i, \eff^{+}) \vee {}\\
& \qquad \Actionclause_{\onep(\overrightarrow{\ob})}(\operator^i, \PARVARS^i, \eff^{-})) \\
\end{align*}

\end{definition}

Consider the positive precondition constraint: in plain words it expresses that if some action occurs at time step $i$,
and some fluent (grounded predicate) is in the positive precondition of that action, then the corresponding variable 
is true at time step $i$. Similarly, the proposition variables corresponding to the positive (negative) effects should be set to true (false) at time step $i+1$. The last constraint encodes the frame axiom: it expresses that either the value of a proposition variable
stays the same, or some positive or negative effect occurs that defines this proposition.

We still need to encode when a positive/negative condition/effect $\Phi$ is associated with the current action.
We also need to define the set of grounded instances of a predicate.

\begin{definition}
\label{def:groundingfunction}
Given an action $\op \in \operator$, a set of atoms $\Phi \in \{\pre^{+}, \pre^{-}, \eff^{+}, \eff^{-}\}$,
and sequence of objects $\overrightarrow{\ob}\in\OBJ^n$ with $n=\arity(\op)$,
its \textbf{grounding}
$\ground(\Phi, \op, \overrightarrow{\ob})\subseteq \St$ is a set of fluents, defined
$\ground(\Phi, \op, \overrightarrow{\ob}):= 
\{ \phi[\overrightarrow{x} / \overrightarrow{\ob}] \mid \phi\in\Phi({\op})\} $.
\end{definition}
The grounding of an action with given object parameters returns the corresponding grounded preconditions/effects.
In the running example, the positive precondition of action $\stack$ for object parameters $(b_1, b_2)$ is grounded as
$\ground(\pre^+, \stack, (b_1, b_2)) = \{\clear(b_1), \clear(b_2), \ontable(b_1)\}$.

Below, we use ``$\bin$'' to express the logarithmic encoding of objects by $\gamma$ bits 
and of action names by $\sigma$ bits. We also use ``='' to denote a bit-wise conjunction of
bi-implications.

\begin{definition}
\label{def:groundedactionclause}
Given the set of action names $\operator$
and a set of atoms $\Phi \in \{\pre^{+}, \pre^{-}, \eff^{+}, \eff^{-}\}$ and a fluent $\onef$,
the \textbf{grounded action constraint} 
$\Actionclause_{\onef}(\operator^i, \PARVARS^i, \Phi)$ =

\begin{align*}
&\bigvee_{\op \in \operator} \hspace{-0.2cm}\bigvee_{\begin{array}{c}\scriptstyle n=\arity(a)\\[-1mm]\scriptstyle\overrightarrow{\ob}\in\OBJ^n\end{array}}\hspace{-0.5cm}
\big\{ ( \overrightarrow{\operator^i} = \bin(\op)) \land
\bigwedge^n_{j=1} (\overrightarrow{\onepm^i_j} = \bin(\ob_j)) \mid
\\[-0.5cm]
& \hspace{5.5cm}\onef \in \ground(\Phi, \op, \overrightarrow{\ob})\big\}
\end{align*}
\end{definition}
In the grounded action constraint, for each grounded action $\op(\overrightarrow{\ob})$ that 
contains fluent $\onef$ in its grounded precondition/effect $\Phi$, we generate equality constraints for action and parameter variables. For example, for fluent $\clear(b_1)$, $\Phi=\pre^+$ and grounded action 
$\stack(b_1,b_2)$, we generate a disjunct 
$\overrightarrow{\operator^i}=\bin(\stack) \wedge \overrightarrow{x_1^i}=\bin(b_1)\wedge \overrightarrow{x_2^i}=\bin(b_2)$.

\smallskip

Finally, due to the use of logarithmic variables, invalid actions are possible when the number of objects and actions are non-powers of $2$.
To maintain consistency we restrict the invalid action and parameter values by the $\RC$-constraint.
Here $<$ denotes the bit-blasted comparison operator on binary numbers.

\begin{definition}
\label{def:restricteclause}
Restricted constraints $\RC(\operator^i, \PARVARS^i)$ =
\begin{displaymath}
\big(\overrightarrow{\operator^i} < |\operator|\big) \land 
\bigwedge_{1 \leq j \leq \maxpar} \big(\overrightarrow{\onepm^i_{j}} < |\OBJ|\big)
\end{displaymath}
\end{definition}

In practice, this constraint can still be refined if one has type information on the objects,
as in typed PDDL domains.

\section{New Ungrounded QBF encoding}
\label{sec:ungroundedqbfencoding}

Similar to the intermediate SAT encoding, the QBF encoding of a plan of length $k$ corresponds to finding a path of length $k$ from the initial state to some goal state.
From the SAT encoding, we can observe that the values of the propositional variables only depend on the action variables, i.e., given a sequence of actions and (complete) initial state, the intermediate states until the goal state are completely determined.
From the constraints on the propositional variables, we can observe that their values are independent of each other,
so these constraints can be enforced independently (at each time step and for each action).
Fluents are generated by grounding the predicates, i.e., by instantiating all object combinations.
The propositional variables are then generated for each fluent and each time step.

For the QBF encoding, the idea is to avoid grounding by representing all object combinations as universal variables and generate constraints on the predicate variables directly.
Essentially the universal object combination variables and existential predicate variables replace the existential propositional variables in the intermediate SAT encoding.

The action variables ($\operator$) and action parameter variables ($\PARVARS$) are the same as for SAT encoding.
We define object combinations and predicates separately.
The object combination variables are $\OBJSVARS = \{ \oneoc_{j,b} \mid 1\leq j \leq \maxarg, 1\leq b\leq \numvar\}$ where $\maxarg$ is the maximum arity of predicates; and $\numvar$ is defined before. Here, $\oneoc_{j,b}$ represents the $b$-th bit of the $j$th object variable and $\oneoc_{j}$ (sequence of $\numvar$ variables) represents the $j$th object variable.
The predicate variables are $\PR^i = \{\oneq^{i}_{\onep} \mid \onep \in \PR\}$ for each $0\leq i \leq k$.
Note that $\operator$ and $\PR$ are ungrounded!

\begin{figure}[p]
\fbox{\parbox{\columnwidth}{
\begin{align*}
&\exists \overrightarrow{\operator^0} \exists \overrightarrow{\onepm^0_{1}} \exists\overrightarrow{\onepm^0_{2}} \phantom{a} \exists \overrightarrow{\operator^1} \exists \overrightarrow{\onepm^1_{1}} \exists\overrightarrow{\onepm^1_{2}} \\
&\forall \overrightarrow{y_1} \forall \overrightarrow{y_2}\\
&\exists \oneq_{\clear}^0 \exists \oneq_{\ontable}^0 \exists \oneq_{\on}^0\\
&\exists \oneq_{\clear}^1 \exists \oneq_{\ontable}^1 \exists \oneq_{\on}^1\\
&\exists \oneq_{\clear}^2 \exists \oneq_{\ontable}^2 \exists \oneq_{\on}^2\\
&\textbf{Initial state: }\\
&((\bin(b_2) = \overrightarrow{y_1}) \iff \oneq_{\clear}^0) \land {}\\
&((\bin(b_1) = \overrightarrow{y_1}) \iff \oneq_{\ontable}^0) \land {} \\
&((\bin(b_2) = \overrightarrow{y_1}) \land (\bin(b_1) = \overrightarrow{y_2}) \iff \oneq_{\on}^0) \land{}\\
&\textbf{Goal state: }\\
&((\bin(b_1) = \overrightarrow{y_1}) \land (\bin(b_2) = \overrightarrow{y_2})) \implies \oneq_{\on}^2 \land{}\\
& \textbf{For time steps i = 0,1}\\
& \textbf{clear:}\\
&\big((\overrightarrow{\operator^i} = \bin(\stack) \land 
	( \overrightarrow{x_1^i} = \overrightarrow{y_1} \lor \overrightarrow{x_2^i} = \overrightarrow{y_1})) \vee {}\\
&\quad\text{ }(\overrightarrow{\operator^i} = \bin(\unstack) \land \overrightarrow{x_1^i} = \overrightarrow{y_1})\big) \implies \oneq_{\clear}^i  \land {}\\
&(\overrightarrow{\operator^i} = \bin(\unstack) \land \overrightarrow{x_2^i} = \overrightarrow{y_1}) \implies \oneq_{\clear}^{i+1} \land \\
&(\overrightarrow{\operator^i} = \bin(\stack) \land \overrightarrow{x_2^i} = \overrightarrow{y_1}) \implies \neg \oneq_{\clear}^{i+1} \land {}\\
& ((\oneq_{\clear}^i = \oneq_{\clear}^{i+1}) \lor (\overrightarrow{\operator^i} = \bin(\unstack) \land \overrightarrow{x_2^i} = \overrightarrow{y_1})\\
&\quad{}\lor(\overrightarrow{\operator^i} = \bin(\stack) \land \overrightarrow{x_2^i} = \overrightarrow{y_1})) \text{ } \land {}\\
& \textbf{ontable:}\\
&(\overrightarrow{\operator^i} = \bin(\stack) \land \overrightarrow{x_1^i} = \overrightarrow{y_1}) \implies \oneq_{\ontable}^{i} \land {}\\
&(\overrightarrow{\operator^i} = \bin(\unstack) \land \overrightarrow{x_1^i} = \overrightarrow{y_1}) \implies \oneq_{\ontable}^{i+1} \land {}\\
&(\overrightarrow{\operator^i} = \bin(\stack) \land \overrightarrow{x_1^i} = \overrightarrow{y_1}) \implies \neg \oneq_{\ontable}^{i+1} \land {}\\
& ((\oneq_{\ontable}^i = \oneq_{\ontable}^{i+1}) \lor (\overrightarrow{\operator^i} = \bin(\unstack) \land \overrightarrow{x_1^i} = \overrightarrow{y_1})\\
& \quad {}\lor(\overrightarrow{\operator^i} = \bin(\stack) \land \overrightarrow{x_1^i} = \overrightarrow{y_1})) \land {}\\
& \textbf{on:}\\
&(\overrightarrow{\operator^i} = \bin(\unstack) \land \overrightarrow{x_1^i} = \overrightarrow{y_1} \land \overrightarrow{x_2^i} = \overrightarrow{y_2}) \implies \oneq_{\on}^{i} \land {}\\
&(\overrightarrow{\operator^i} = \bin(\stack) \land \overrightarrow{x_1^i} = \overrightarrow{y_1} \land \overrightarrow{x_2^i} = \overrightarrow{y_2}) \implies \oneq_{\on}^{i+1} \land {}\\
&(\overrightarrow{\operator^i} = \bin(\unstack) \land \overrightarrow{x_1^i} = \overrightarrow{y_1} \land \overrightarrow{x_2^i} = \overrightarrow{y_2}) \implies \neg \oneq_{\on}^{i+1} \land {}\\
& ((\oneq_{\on}^i = \oneq_{\on}^{i+1}) \lor (\overrightarrow{\operator^i} = \bin(\stack) \land \overrightarrow{x_1^i} = \overrightarrow{y_1} \land \overrightarrow{x_2^i} = \overrightarrow{y_2})\\
&\quad{}\lor(\overrightarrow{\operator^i} = \bin(\unstack) \land \overrightarrow{x_1^i} = \overrightarrow{y_1} \land \overrightarrow{x_2^i} = \overrightarrow{y_2}))
\end{align*}}}
\caption{Ungrounded Encoding for the blocks-world domain with a path length of 2 actions}
\label{fig:qbfex}
\end{figure}

\medskip
\noindent The corresponding Ungrounded Encoding is:
\begin{align*}
&\exists \operator^0, \PARVARS^0, \dots, \operator^{k-1}, \PARVARS^{k-1} \forall \OBJSVARS \exists \PR^0, \dots, \PR^k \\
&\UI(\PR^0, \OBJSVARS) \land \UG(\PR^k, \OBJSVARS) \land {}\\
&\bigwedge_{i=0}^{k-1}\UTi(\PR^i, \PR^{i+1}, \OBJSVARS, \operator^i, \PARVARS^i) \land \bigwedge_{i=0}^{k-1} \RC(\operator^i, \PARVARS^i)
\end{align*}

In the sequel, we will define the auxiliary constraints for the initial and goal states and the transition functions. The domain restriction $\RC$ does not depend on the universal variables $\OBJSVARS$, so it remains unchanged from the SAT encoding (refer Definition~\ref{def:restricteclause}). 
We refer to the example in 
Figure~\ref{fig:qbfex} for an illustration of the encoding in a concrete example. Finally, we comment on the equivalence
of the SAT and the QBF encoding.
When the universal variables in the ungrounded QBF encoding expanded, the resulted conjunction of constraints have correponding equivalent constraints in the intermediate SAT encoding.

 The initial and goal constraints must be consistent with object combinations, so for each predicate, we specify in which forall branches, i.e., for which objects instantiations it is positive or negative.
\begin{definition}
The \textbf{Ungrounded Initial} constraint $\UI(\PR^0, \OBJSVARS)$:
\begin{displaymath}
\bigwedge_{\onep \in \PR} \big(\bigvee_{\onep(\overrightarrow{\ob}) \in \initial} \bigwedge^{|\overrightarrow{\ob}|}_{j=1} 
(\bin(\ob_j) = \overrightarrow{\oneoc_j})\big) \iff \oneq^{0}_{\onep}
\end{displaymath}
\end{definition}

\begin{definition}
The \textbf{Ungrounded goal} constraint $\goal_u(\PR^{k}, \OBJSVARS) = \goal^{+} \land \goal^{-}$ where
\begin{align*}
&\goal^{+} = \bigwedge_{\onep \in \PR}\big( \bigvee_{\onep(\overrightarrow{\ob}) \in \oneg^{+}} 
\bigwedge^{|\overrightarrow{\ob}|}_{j=1} (\bin(\ob_j) = \overrightarrow{\oneoc_j})\big) \implies \oneq_{\onep}^{k},\\
&\goal^{-} = \bigwedge_{\onep \in \PR}\big( \bigvee_{\onep(\overrightarrow{\ob}) \in \oneg^{-}} 
\bigwedge^{|\overrightarrow{\ob}|}_{j=1} (\bin(\ob_j) = \overrightarrow{\oneoc_j})\big) \implies \neg \oneq_{\onep}^{k}
\end{align*}
\end{definition}

\begin{definition}
\label{def:ungroundedtf}
The \textbf{ungrounded transition function}
\begin{displaymath}
\UTi(\PR^i, \PR^{i+1}, \OBJSVARS, \operator^i, \PARVARS^i) = {\bigwedge_{\onep \in \PR} \Uclause_{\onep}^{i}}
\end{displaymath}
where the \textbf{ungrounded predicate constraint} $\Uclause^{i}_{\onep}$ =
\begin{align*}
&(\Actionclause_{\onep}^{u}(\operator^i, \PARVARS^i, \OBJSVARS, \pre^{+}) \implies \oneq^i_{\onep}) \wedge {}\\
&(\Actionclause_{\onep}^{u}(\operator^i, \PARVARS^i, \OBJSVARS, \pre^{-}) \implies \neg \oneq^i_{\onep}) \wedge {}\\
&(\Actionclause_{\onep}^{u}(\operator^i, \PARVARS^i, \OBJSVARS, \eff^{+}) \implies \oneq^{i+1}_{\onep}) \wedge {}\\
&(\Actionclause_{\onep}^{u}(\operator^i, \PARVARS^i, \OBJSVARS, \eff^{-}) \implies \neg \oneq^{i+1}_{\onep}) \wedge {}\\
&((\oneq^{i}_{\onep} = \oneq^{i+1}_{\onep}) \vee \Actionclause_{\onep}^{u}(\operator^i, \PARVARS^i, \OBJSVARS, \eff^{+}) \vee {} \\
&\qquad \Actionclause_{\onep}^{u}(\operator^i, \PARVARS^i, \OBJSVARS, \eff^{-})) \\
\end{align*}

\end{definition}

For each predicate, if any of the action constraints is true in the corresponding preconditions or effects, then the corresponding variable is constrained to the appropriate value.
We will expand on the action constraints in the following definition.
The constraints are similar to that of SAT encoding, for example the positive precondition constraint represents if the action with the predicate as positive precondition is true then the corresponding predicate variable is set to true in time step $i$.
The last constraint is called the frame axiom; it says that either the value of predicate stays the same or some positive or negative effect constraint is true.

\begin{definition}
\label{def:ungroundedactionclause}
Given the set of actions $\operator$ and $\Phi \in \{\pre^{+}, \pre^{-}, \eff^{+}, \eff^{-}\}$, the \textbf{ungrounded action constraint}
\begin{align*}
& \Actionclause_{\onep}^{u}(\operator^i, \PARVARS^i, \OBJSVARS, \Phi) =
\bigvee_{\op \in \operator}  (\overrightarrow{\operator^i} = \bin(\op)) \land {}\\
&\qquad \bigvee \big\{\bigwedge^{n}_{j=1} 
(\overrightarrow{\onepm^{i}_{l_j}} = \overrightarrow{\oneoc_j}) \mid
\onep(\onepm_{l_1}, \dots, \onepm_{l_n} ) \in \Phi(\op)\big\}
\end{align*}

\end{definition}

Similar to the initial and goal constraints, the value of the predicate variables must be constrained in each forall branch.
As we discuss before, each forall branch corresponds to one instantiation of object combinations.
The ungrounded action constraint specifies which branches are evaluated to true.
In plain words, for every forall branch where the predicate parameters variables are equal to the universal variables the ungrounded action constraint is evaluated to true.
Note that the predicate parameters from the preconditions and effects are simply a subset of the action parameter variables.

Regarding the frame axioms, we use the following performance optimisations in our implementation:
We do not need to propagate static predicates, i.e., predicates that do not appear in any of the action effects.
Instead, for static predicates we introduce only one copy, used across all time steps.
In particular, the type predicates in typed planning domains are handled as static predicates.
Equality-predicates are treated specially, since they impose constraints between action parameters only, 
independent of forall variables.
We simply generate (in)equality constraints between action parameters directly, similar to the RC-constraint.

\subsection{Correctness}
\label{sec:correctness}

We demonstrate correctness, by showing how the SAT encoding can be stepwise transformed into the equivalent QBF encoding.
Let $\numoc = 2^{\numvar \times \maxarg}$ be the number of object combinations, and $n = |\PR|$. Then the constraints on the propositions in time step $i$ and $i+1$ in the SAT encoding are:

\begin{align*}
&\exists \onef^{i}_{\onep_{1}(\overrightarrow{\ob_{1}})} \dots \exists \onef^{i}_{\onep_{1}(\overrightarrow{\ob_{\numoc}})}\\
& \dots\\
&\exists \onef^{i}_{\onep_{n}(\overrightarrow{\ob_{1}})} \dots \exists \onef^{i}_{\onep_{n}(\overrightarrow{\ob_{\numoc}})}\\
&\qquad\bigwedge_{\onep \in \PR}\bigwedge_{\onep(\overrightarrow{\ob}) \in \St} \Clause_{\onep(\overrightarrow{\ob})}^i
\end{align*}
Here the existential blocks of propositional variables are grouped based on the predicate symbols.
Since each of the propositions appears in only one constraint, we can push the existential variables inside the conjuction.
Thus the constraints are equivalent to the following groups of constraints in non-prenex form:
\begin{align*}
&\exists \onef^{i}_{\onep_{1}(\overrightarrow{\ob_{1}})} \dots \exists \onef^{i}_{\onep_{1}(\overrightarrow{\ob_{\numoc}})} \bigwedge_{\onep_{1}(\overrightarrow{\ob}) \in \St} \Clause_{\onep_{1}(\overrightarrow{\ob})}^i \\
& {} \land \dots \land {}\\
&\exists \onef^{i}_{\onep_{n}(\overrightarrow{\ob_{1}})} \dots \exists \onef^{i}_{\onep_{n}(\overrightarrow{\ob_{\numoc}})} \bigwedge_{\onep_{n}(\overrightarrow{\ob}) \in \St} \Clause_{\onep_{n}(\overrightarrow{\ob})}^i
\end{align*}

Since all the constraints have a uniform shape, we can abbreviate the conjunction by a for-all quantification
in QBF. So the previous formula is equivalent to the grouping with object combinations variables in the QBF encoding:
\begin{align*}
&\forall \OBJSVARS \text{ } \exists \oneq^{i}_{\onep_{1}} \Uclause_{\onep_{1}}^i \land \dots \land \exists \oneq^{i}_{\onep_{n}} \Uclause_{\onep_{n}}^i
\end{align*}

Remember that the action and parameter variables are the same for both SAT and QBF encoding.
This proves the equivalence between the SAT and the QBF encodings.

\begin{table*}
\centering
\begin{tabular}{lccccccccccccccc}
\toprule
\phantom{abc} &\multicolumn{3}{c}{Q-planner} & \phantom{a} & \multicolumn{3}{c}{Madagascar} & \phantom{a} &\multicolumn{3}{c}{FDS} & \phantom{a} &\multicolumn{3}{c}{PL}\\
\cmidrule(r){2-4} \cmidrule(r){6-8} \cmidrule(r){10-12} \cmidrule(r){14-16}
Domains (Total) & I & UI & pm & & I & UI & pm & & I & UI & pm & & I & UI & pm\\
\midrule
OS-Sat18 ($20$) & $12$ & $1$ & $4.9$ & & - & - & - & & $3$ & - & $2.2$ & & $\mathbf{15}$ & $\mathbf{4}$ & $1.1$ \\
OS-Opt18 ($20$) & $\mathbf{20}$& -  & $2.6$ & & - & - & - & & $9$ & - & $15.7$ & & $\mathbf{20}$ & - & $8.4$ \\
Alkene ($18$) & $\mathbf{18}$ & - & $0.06$ & & $1$ & - & $115.9$ & & $\mathbf{18}$ & - & $15.7$ & & $\mathbf{18}$ & - & $0.02$ \\
MitExams ($20$) & $8$ & $1$ & $4.8$ & & - & - & - & & $1$ & - & $1.8$ & & $\mathbf{12}$ & $\mathbf{5}$ & $8.4$ \\
\bottomrule
\end{tabular}
\caption*{Table $1$: Running 4 tools with $300$ GB memory and $3$ hour time limit on $4$ Organic-Synthesis domains.
We report ``I'', the number of solved instances, ``UI'', the number of uniquely solved instances and ``pm'', the peak memory for the solved instances in GB.\vspace{-2mm}}
\label{table:completeosdata}
\end{table*}

\section{Implementation and Experiments}
\label{sec:analysis}
We implemented our encoding of PDDL problems to QBF problems in QCIR format as a Python program,
which is available online.\footnote{https://github.com/irfansha/Q-Planner} 
The tool handles domains with types and equality predicates by treating them as static predicates;
it also handles negative preconditions and constants, but it cannot yet handle domains with
conditional effects.
To solve the encoded problems for a given length $k$, we transform them to QDIMACS format,
and use the QBF solver CAQE \cite{DBLP:conf/fmcad/RabeT15} 
with the internal preprocessor Bloqqer \cite{DBLP:conf/cade/BiereLS11}. 
Some initial experiments with other preprocessors and solvers indicated that CAQE+Bloqqer is a very
good combination; we leave a systematic comparison as future work. 
Finally, our tool extracts a concrete plan from CAQE's output, and validates that the plan is indeed correct. 
In the sequel, we refer to the combination of our encoding, CAQE and Bloqqer as ``Q-Planner''.

We performed two experiments: in the first experiment, we run Q-Planner on planning problems from 4
Organic Synthesis domains: $2$ from IPC-18 (non-split Satisfying and Optimizing track) and the 2 original benchmarks 
(Alkene and Mitexams) \cite{DBLP:conf/gcai/MasoumiAS15}.\footnote{https://www.cs.ryerson.ca/mes/publications/}
The domains submitted to IPC-18 are simplified domains of the original benchmarks, which received the outstanding domain submission award. We compare Q-Planner with 3 state-of-the-art planners, specified below.

In the second experiment, we encoded all classical planning problems from previous IPC planning
competions that our translation can handle, resulting in 20 domains, including 2 for Organic Synthesis.
We are mainly interested in the number
of instances that can be solved in each domain (within a given time and memory limit).
In this experiment, we compared Q-Planner with Madagascar, using
a simple SAT encoding (no invariants or parallel plans) to compare the effect of our ungrounded
encoding on the encoding size, memory usage and solving time.

\subsection{Experimental Setup}
\label{subsec:experimentalsetup}

For the first experiment, we compare Q-planner with 3 other planners: (1) Madagascar \cite{Rintanen2014MadagascarS} with relaxed existential step encoding \cite{RINTANEN20061031}, a SAT based planner which first grounds the planning problem; (2) Fast Downward Stone Soup 2018 \cite{Helmert2011FastDS}, a non-SAT based planner which was the winner of IPC-2018 competitions in the satisfying track; and (3) Powerlifted (PL) \cite{correa2020lifted}, a non-SAT based planner which avoids grounding and is the state-of-the-art for the Organic synthesis domains.
We use recommended configurations for all three planners for fair comparisons.
We allow $300$GB of main memory and $3$ hours of time limit for solving each instance
of Organic Synthesis.

For the second experiment, we want to compare the ungrounded QBF encoding (using Q-Planner) 
with the grounded SAT encoding (using Madagascar). 
To eliminate other factors, we use Madagascar in its simplest configuration without invariants and with sequential plan
semantics (here called M-simple),
i.e., a standard sequential SAT encoding with direct encoding of objects and actions.
Since the CAQE solver calls SAT-solver Crypto-minisat \cite{DBLP:conf/sat/SoosNC09}, we also use Crypto-minisat to solve the SAT encodings in Madagascar.

For each instance, we generate encodings of increasing plan length in steps of $5$ (steps of $1$ for Organic Synthesis domains) until a solution is found. We abort the run when either a time limit of $5000$ seconds is
reached or the peak memory exceeds $8$GB.
We iterate over all instances within a given domain, but we stop once there are $5$ timed-out instances since we assume that their difficulty increases (mostly due to an increase in the number of objects).
Since Bloqqer is called internally by CAQE, the solving time includes the preprocessing time.

We ran our experiments on a 
Grendel-s\footnote{http://www.cscaa.dk/grendel-s/} 
Huawei FusionServer Pro V1288H V5 server nodes for all the computations, each with 48 cores of 3.0 GHz (Intel Xeon Gold 6248R) and 384 GB main memory (but only one core is used per problem).

\subsection{Results}
\label{subsec:results}

\begin{table}[t]
\centering
\begin{tabular}{cl|cc|cc}
\toprule
& Domains & \multicolumn{2}{c|}{Q-planner} &  \multicolumn{2}{c}{M-simple}\\
& & \#solved & size & \#solved & size \\
\midrule
\parbox[t]{2mm}{\multirow{11}{*}{\rotatebox[origin=c]{90}{Typed IPC domains}}}
&Blocks & $28$ & 0.14 & $\mathbf{30}$ & 4.8 \\
&DriverLog & $\mathbf{11}$ & 0.27 & $\mathbf{11}$ & 5.7\\
&Elevator & $34$ & 0.20 & $\mathbf{43}$ & 2.1 \\
&Hiking & $\mathbf{9}$ & 0.49 & $6$ & 75.8 \\
&OS-Opt18 & $\mathbf{20}$ & 2.6 & $0$ & - \\
&OS-Sat18 & $\mathbf{11}$ & 3.2 & $0$ & - \\
&Satellite & $\mathbf{6}$ & 0.23 & $\mathbf{6}$ & 1.7 \\
&Tidybot & $1$ & 1.7 & $\mathbf{13}$ & 519.9 \\
&Termes & $2$ & 0.68 & $\mathbf{3}$ & 14.3 \\
&Thoughtful & $0$ & 1.7 & $\mathbf{5}$ & 857.5\\
&Visitall & $7$ & 0.41  & $\mathbf{13}$ & 2.5 \\
\midrule
\parbox[t]{2mm}{\multirow{9}{*}{\rotatebox[origin=c]{90}{Untyped IPC domains}}}
&Blocks-3op & $17$ & 0.16 & $\mathbf{19}$ & 63.2 \\
&Movie & $\mathbf{30}$ & 0.15 & $\mathbf{30}$ & 0.28\\
&Depot & $4$ & 0.32 & $\mathbf{6}$ & 70.9 \\
&Gripper & $3$ & 0.19 & $\mathbf{5}$ & 0.80 \\
&Logistics & $\mathbf{12}$ & 0.34 & $\mathbf{12}$ & 4.0 \\
&Mprime & $17$ & 0.36 & $\mathbf{34}$ & 1954 \\
&Mystery & $3$ & 0.36 & $\mathbf{13}$ & 399.1 \\
&Grid & $\mathbf{2}$ & 0.52 & $\mathbf{2}$ & 47.9\\
&Freecell & $2$ & 1.1 & $\mathbf{6}$ & 631.1 \\
\midrule
&Total sum & 219 & 15.12 & 257 & 4656 \\
\bottomrule
\end{tabular}
\caption*{Table $2$: Instances solved within $5000$s and $8$GB
internal memory.
We also show the size (in MB) of the encoding for the maximum instance after preprocessing (if possible).
\vspace{-0.3cm}}
\label{table:domains}
\end{table}

For the first experiment (Table $1$), Q-Planner solves $58$ instances in total, while never using more than $4.9$GB memory for all the solved instances; whereas Madagascar only solves $1$ instance using $116$GB memory.
These results are consistent with the IPC-$2018$ competitions, where $2$ variants of Madagascar could not solve any of the (non-split) organic synthesis benchmarks in the satisfying track based on SAT solving.
This is due to the presence of actions with many parameters ($>17$), thus even with action splitting and other optimisations the grounded encodings are too big.
It is clear that Q-Planner outperforms Madagascar significantly.

Concerning the non-SAT based planners: FDS, which has a special strategy for partial grounding, solves $31$ instances in total, but on the other instances
it still exhausts all the memory during the grounding (translation) phase.
This emphasizes the problem with grounding even in non-SAT based planners.
The Powerlifted planner solves $65$ instances, in total solving $7$ more instances than Q-planner.
Interestingly, Q-Planner solves one instance uniquely in the hardest set of benchmarks, i.e., MitExams.
The good performance of Powerlifted is expected, since it also avoids grounding on actions while using optimized database queries on grounded states.
We expect this is due to the large number of grounded propositions in the initial state which is harder for Powerlifted.

We found some interesting observations regarding the Powerlifted planner:
If it finds a plan successfully, then it never takes more than a couple of hundred MB and finds a plan quickly 
(except for one instance in both MitExams and OS-Opt18, which takes $8.4$GB). 
On the failed instances, however, Powerlifted either exhausts all the memory or runs out of time.
Our Q-Planner, on the other hand, might exceed the time-limit, but never used more than $20$GB for refuting the existence of a plan of shorter lengths, even when it could not solve the instance completely.
For the unsolved instances in domains MitExams and OS-Sat18, the minimum and maximum plan refutation length is $3$ and $7$, respectively, with an average length of $5$.
This is the strength of SAT/QBF based planners, for problems where quickly refuting the existence of a plan 
of bounded length is useful.

\medskip
Concerning the second experiment, the Ungrounded QBF Encoding never exceeds $2$MB (or $3$MB in case of Organic Synthesis),
so it is orders of magnitude more compact than the SAT encoding (which sometimes exceeds $1950$MB).
When comparing the time for solving, the situation is different: Madagascar performs better than Q-Planner on most other benchmarks (except Hiking), with better peak memory and running times.
This is expected since most benchmarks in the IPC domains are easy to ground.
For domains with a large number of action parameters (Hiking, Organic Synthesis), our QBF
encoding outperforms the SAT encoding.

\begin{figure}[t]
\centering
\includegraphics[width=0.9\columnwidth]{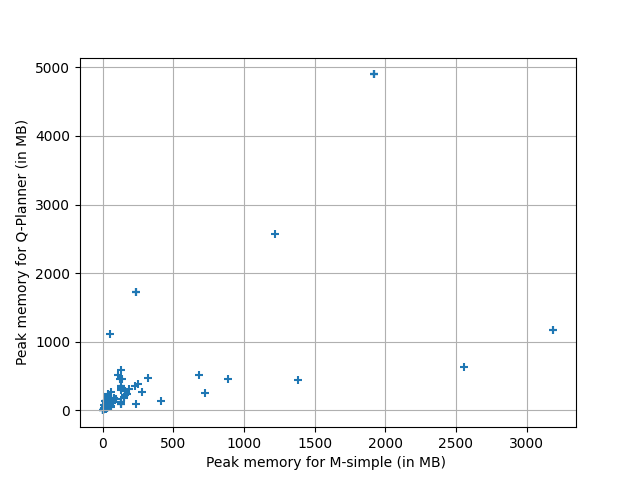}
\caption{Peak memory of M-simple and Q-planner}.
\label{fig:peakmemory}
\end{figure}

The peak memory (see Figure \ref{fig:peakmemory}) consumed by solving the Ungrounded Encodings is usually higher than the SAT encodings. This reflects the fact that QBF is PSPACE complete, where SAT is merely NP-complete.
In Figure \ref{fig:peakmemory},
only instances which are solved by both encodings are compared. However, there are some instances where UE takes less memory than the SAT encoding, and the UE encoding solves the problem whereas the SAT encoding could not (Hiking and Organic Synthesis benchmarks).
The higher peak memory usage by the CAQE solver can be understood by considering its solution approach.
CAQE solver generates abstract propositional clauses and calls SAT solver (Crypto-minisat) as a subroutine while taking advantage of the structure of the QBF formula given.
While the peak memory is higher for CAQE, the structure of the QBF encoding allows it to handle the grounding bottleneck which we will see in the Organic Synthesis problems.
We leave experiments with other QBF solvers for future work.

\begin{figure}[t]
\centering
\includegraphics[width=0.9\columnwidth]{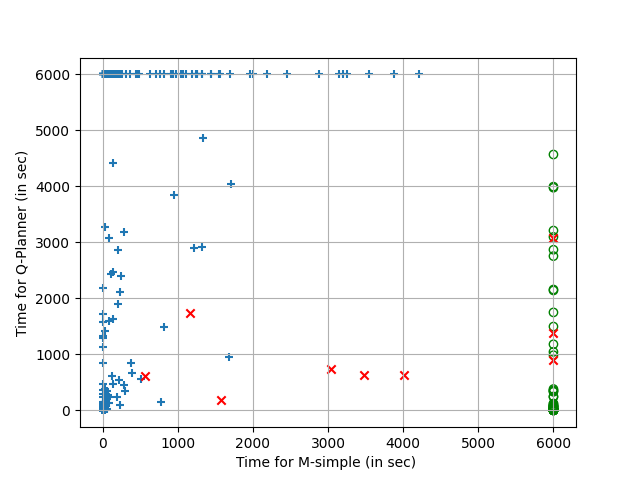}
\caption{Solving times of M-simple and Q-Planner}.
\label{fig:solvingtimes}
\end{figure}

From the results on $20$ IPC domains, it is clear that (with current QBF solvers) our QBF encoding  is not a replacement for SAT based solvers, but rather a complement for domains where grounding is hard due
to many action parameters.

\section{Related Work}
\label{sec:relatedwork}

We first discuss some approaches in the literature that avoid grounding in SAT encodings 
by allowing quantification, using EPR, CP, or SMT. Then we discuss related work on QBF.

\citet{DBLP:conf/birthday/PerezV13} presented an encoding of planning problems in EPR (Effectively Propositional Logic), a fragment of First Order Logic.
While EPR can encode planning problems with fluent dependencies, the satisfiability problem of this class is NEXPTIME-complete~\cite{EPR}.
Because we use only one universal block of variable branches, our QBF encoding still stays in the 3rd level of the polynomial hierarchy, combining a compact encoding with relatively efficient solving.

Other approaches, such as numerical planning as SMT (Satisfiability Modulo Theories) by
\citet{DBLP:journals/ker/BofillEV16} and numerical planning as CP (Constraint Programming) by  \citet{espasa2019towards}, also avoid grounding for compact encodings.
It would be interesting to apply those approaches to Organic Synthesis.

\citet{cashmore2013partially} proposed a QBF encoding with \emph{partial grounding}, which grounds 
a subset of the propositions and handles the ungrounded propositions using universal variables.
Our work extends their partially grounded QBF encoding, whose core idea uses
predicate variables to represent ungrounded propositions and
\emph{universal variables for only one object}.
Equation \ref{eq:partial} from \citet{cashmore2013partially}, shows 
the prefix for their proposed encoding.
In plain words, it states that there exists an assignment to the grounded variables and action-lock variables (i.e., ungrounded parameter variables) such that for all assignments of one object, there exists an assignment to $n+1$ sets of predicate variables, such that the constraints hold.
\begin{equation}\label{eq:partial}
\exists X^{g}_{1} \dots X^{g}_{n+1} \forall a_1 \dots a_m \exists X^{u}_{1} \dots X^{u}_{n+1}
\end{equation}
However, the problem with their encoding comes with using universal variables to represent only one object.
However, this leads to problems with propagating untouched propositions at each time step.
By quantifying over one object only, \emph{not all} values of propositions are propagated.
To avoid incorrect plans, they must resort to \emph{partial} grounding: the restriction
is that every predicate can have at most one ungrounded parameter. This can become costly:
When a predicate parameter is grounded, all the corresponding action parameters need to be grounded as well,
resulting in a large number of partially grounded actions.

We see our work as a generalisation of the technique by \citet{cashmore2013partially}. We consider 
\emph{object combinations} in the universal layer of the encoding, thus avoiding grounding completely.
This results in more compact QBF encodings.
Since the software from partial grounding encoding is not available to us we could not compare it with Q-planner.
Partial grounding of some parameters is trivial in our approach (it can be done on the domain file directly), however, we expect that the result will be suboptimal.
In case of organic synthesis, partially grounding some predicates still results in very large encodings.

The main focus of compact QBF encodings for reachability, synthesis and planning has been on reducing the number of transition function copies \cite{DBLP:conf/lpar/Rintanen01, cashmore2012planning}.
While avoiding unrolling reduces the encoding size, it comes at the cost of losing inference between different time steps.
On the other hand, using universal quantification to avoid grounding, while keeping the inference between time steps,
might be a better way to use the advances in the QBF solving, as shown for planning problems in this paper.
Since the encodings are generated directly from (restricted) First-order Logic representations, the resulting encodings are compact and more scalable.

\section{Conclusion and Future Work}
\label{sec:conclusionandfuturework}

In this paper, we propose the Ungrounded Encoding for classical planning problems to QBF,
which results in compact encodings by avoiding grounding completely.
The size of this QBF encoding is linear in the number of action names, predicates
and the path length, and logarithmic in the number of objects.
We provide an open source implementation of the Ungrounded Encoding, translating
classical planning problems in PDDL to QBF problems in QCIR format. 
The resulting QBF formulas can be preprocessed and solved by
existing QBF tools (we used Bloqqer and CAQE).

The experiments show that our encodings effectively avoid the memory bottleneck 
due to grounding. This is relevant for domains with actions 
that involve many objects, such as Organic Synthesis.
Our main result is to solve many instances of Organic Synthesis, which so far could not be handled by any SAT/QBF based planner. We solve $58$ problems in the $4$ Organic Synthesis domains.
We compare our implementation with the state-of-the-art SAT based planner Madagascar, 
which only solves $1$ instance, using $116$GB memory.
We also compare with the non-SAT based planners Fast Downward Stone Soup (FDS-2018) and Powerlifted.
FDS was the winner of the satisfying track in IPC 2018 and solves $31$ instances. Powerlifted avoids grounding actions, and solves $65$ instances altogether.

\paragraph{Future Work.}
There are several research directions from here. One direction is to extend the Ungrounded Encoding to planning problems with uncertainty or planning with adversaries. 
QBF encodings have been demonstrated to be useful for such domains, such as conformant planning
\cite{DBLP:journals/jair/Rintanen99,rintanen2007AOEQBF}. 
To this end, the Ungrounded Encoding must be
extended, in order to handle some dependencies between state predicates, and to handle conditional effects. 

Another direction is to improve the efficiency of the Ungrounded Encoding:
One could consider to incorporate automatically generated invariants \cite{rintanen2008regression} to speed up the search. Similarly, one could consider extending the Ungrounded Encoding
with compact QBF encodings, such as the non-copying Iterative Squaring Encoding \cite{DBLP:conf/lpar/Rintanen01} and the Compact Tree Encoding \cite{cashmore2012planning}. These encodings could
in principle lead to an even more concise encoding (logarithmic in the length of the plan), but we expect
less spectacular benefits than compared to SAT encodings, since our encoding is already quite small.
It would also be interesting to investigate ungrounded versions of \emph{parallel plans} \cite{RINTANEN20061031}, but currently the sequentiality of plans is deeply embedded in our encoding. 

Finally, we hope to inspire new research on pre-processing and solving QBF problems,
by submitting our problems to the QBFeval competition.
Our encodings are very concise, but the current QBF solvers do not yet take advantage of the structure 
of the Ungrounded Encoding. 
While SAT solvers are rather mature, the field of QBF solvers is still improving rapidly.
An advance in QBF solvers could make the Q-Planner approach competitive to the SAT approach
for general classical planning problems.

\subsection*{Acknowledgments}
We thank Michael Cashmore for discussions and clarifications on partially grounded QBF encoding, Leander Tentrup for support regarding the CAQE QBF solver, and Mikhail Soutchanski for providing original Organic Synthesis benchmarks. The numerical  results presented in this work were obtained at the Centre for Scientific Computing, Aarhus \href{http://phys.au.dk/forskning/cscaa/}{(CSCAA)}.

% Use \bibliography{yourbibfile} instead or the References section will not appear in your paper
\bibliography{references}

\end{document}